# Performance Analysis of AIM-K-means & K-means in Quality Cluster Generation

Samarjeet Borah, Mrinal Kanti Ghose

**Abstract**: Among all the partition based clustering algorithms K-means is the most popular and well known method. It generally shows impressive results even in considerably large data sets. The computational complexity of K-means does not suffer from the size of the data set. The main disadvantage faced in performing this clustering is that the selection of initial means. If the user does not have adequate knowledge about the data set, it may lead to erroneous results. The algorithm Automatic Initialization of Means (AIM), which is an extension to K-means, has been proposed to overcome the problem of initial mean generation. In this paper an attempt has been made to compare the performance of the algorithms through implementation

—————————— ◆ ——————————

## 1 INTRODUCTION

Clustering [2][3][4] is a type of unsupervised learning method in which a set of elements is separated into homogeneous groups. It seeks to discover groups, or clusters, of similar objects. Generally, patterns within a valid cluster are more similar to each other than they are to a pattern belonging to a different cluster. The similarity between objects is often determined using distance measures over the various dimensions in the dataset. The variety of techniques for representing data, measuring similarity between data elements, and grouping data elements has produced a rich and often confusing assortment of clustering methods. Clustering is useful in several exploratory pattern-analysis, grouping, decision-making, and machine-learning situations, including data mining, document retrieval, image segmentation, and pattern classification [5][3].

## 2 PARTITION BASED CLUSTERING METHODS

Partition based clustering methods create the clusters in one step. Only one set of clusters is created, although several different sets of clusters may be created internally within the various algorithms. Since only one set of clusters is output, the users must input the desired number of clusters. Given a database of n objects, a partition based [5] clustering algorithm constructs k partitions of the data, so that an objective function is optimized. In these clustering methods some metric or criterion function is used to determine the goodness of any proposed solution. This measure of quality could be average distance between clusters or some other metric. One common measure of such kind is the squired error metric, which measures the squired distance from each point to the centroid for the associated cluster. Partition based clustering algorithms try to locally improve a certain criterion. The majority of them could be considered as greedy algorithms, i.e., algorithms that at each step choose the best solution and may not lead to optimal results in the end. The best solution at each step is the placement of a certain object in the cluster for which the representative point is nearest to the object. This family of clustering algorithms includes the first ones that appeared in the Data Mining Community. The most commonly used are K-means [JD88, KR90][6], PAM (Partitioning Around Medoids) [KR90], CLARA (Clustering LARge Applications) [KR90] and CLARANS (Clustering LARge ApplicatioNS ) [NH94]. All ofthem are applicable to data sets with numerical attributes.

### 2.1 K-means Algorithm

K-means [7] is a prototype-based, simple partitional clustering technique which attempts to find a user-specified *K* number of clusters. These clusters are represented by their centroids. A cluster centroid is typically the mean of the points in the cluster. This algorithm is a simple iterative clustering algorithm. The algorithm is simple to implement and run, relatively fast, easy to adapt, and common in practice. It is historically one of the most important algorithms in data mining. The general algorithm was introduced by Cox (1957), and (Ball and Hall, 1967; MacQueen, 1967) [6] first named it K-means. Since then it has become widely popular and is classified as a *partitional* or *non-hierarchical* clustering method (Jain and Dubes, 1988). It has a number of variations [8][11].

The *K-means* algorithm works as follows:
  a. Select initial centres of the *K* clusters. Repeat steps b through c until the cluster membership stabilizes.
  b. Generate a new partition by assigning each data to its closest cluster centres.
  c. Compute new cluster centres as the centroids of the clusters.

The algorithm can be briefly described as follows:
Let us consider a dataset $D$ having $n$ data points $x_1, x_2$…

- *Samarjeet Borah is with the Department of Computer Science & Engineering, Sikkim Manipal Institute of Technology, Majitar, Rangpo, East Sikkim-737132.*
- *Mrinal Kanti Ghose is with the Department of Computer Science & Engineering as Professor & HOD, Sikkim Manipal Institute of Technology, Majitar, Rangpo, East Sikkim-737132.*

$x_n$. The problem is to find minimum variance clusters from the dataset. The objects have to be grouped into $k$ clusters finding $k$ points $\{m_j\}(j=1, 2, …, k)$ in $D$ such that

$$\frac{1}{n}\sum_{i=1}^{n} \min_j d^2(x_i, m_j) \quad (1)$$

is minimized, where $d(x_i, m_j)$ denotes the Euclidean distance between $x_i$ and $m_j$. The points $\{m_j\}$ $(j=1, 2, …, k)$ are known as cluster centroids. The problem in Eq.(1) is to find $k$ cluster centroids, such that the average squared Euclidean distance (mean squared error) between a data point and its nearest cluster centroid is minimized.

The *K-means* algorithm provides an easy method to implement approximate solution to Eq.(1). The reasons for the popularity of *K-means* are ease and simplicity of implementation, scalability, speed of convergence and adaptability to sparse data. The *K-means* algorithm can be thought of as a gradient descent procedure, which begins at starting cluster centroids, and iteratively updates these centroids to decrease the objective function in Eq.(1). The *K-means* always converge to a local minimum. The particular local minimum found depends on the starting cluster centroids. The *K-means* algorithm updates cluster centroids till local minimum is found. Before the *K-means* algorithm converges, distance and centroid calculations are done while loops are executed a number of times, say $l$, where the positive integer $l$ is known as the number of *K-means* iterations. The precise value of $l$ varies depending on the initial starting cluster centroids even on the same dataset. So the computational time complexity of the algorithm is $O(nkl)$, where $n$ is the total number of objects in the dataset, $k$ is the required number of clusters we identified and $l$ is the number of iterations, $k \leq n$, $l \leq n$.

K-mean clustering algorithm is also facing a number of drawbacks. When the numbers of data are not so many, initial grouping will determine the cluster significantly. Again the number of cluster, K, must be determined before hand. It is sensitive to the initial condition. Different initial conditions may produce different results of cluster. The algorithm may be trapped in the local optimum. Weakness of arithmetic mean is not robust to outliers. Very far data from the centroid may pull the centroid away from the real one. Here the result is of circular cluster shaped because based on distance.

The major problem faced during K-means clustering is the efficient selection of means. It is quite difficult to predict the number of clusters k in prior. The k varies from user to user. As a result, the clusters formed may not be up to mark. The finding out of exactly how many clusters will have to be formed is a quite difficult task. To perform it efficiently the user must have detailed knowledge of the domain. Again the detail knowledge of the source data is also required.

## 2.2 Automatic Initialization of Means

The Automatic Initialization of Means (AIM) [12] has been proposed to make the K-means algorithm a bit more efficient. The algorithm is able to detect the number of total number of clusters automatically. This algorithm also has made the selection process of the initial set of means automatic. AIM applies a simple statistical process which selects the set of initial means automatically based on the dataset. The output of this algorithm can be applied to the K-means algorithm as one of the inputs.

### 2.2.1 Background

In probability theory and statistics, the Gaussian distribution is a continuous probability distribution that describes data that clusters around a mean or average. Assuming Gaussian distribution it is known that $\mu \pm 1\sigma$ contain 67.5% of the population and thus significant values concentrate around the cluster mean $\mu$. Points beyond this may have tendency of belonging to other clusters. We could have taken $\mu \pm 2\sigma$ instead of $\mu \pm 1\sigma$, but problem with $\mu \pm 2\sigma$ is that it will cover about 95% of the population and as a result it may lead to improper clustering. Some points that are not so relevant to the cluster may also be included in the cluster.

### 2.2.2 Description

Let us assume that data set $D$ as $\{x_i, i=1, 2… N\}$ which consists of $N$ data objects $x_1, x_2, …, x_N$, where each object has $M$ different attribute values corresponding to the $M$ different attributes. The value of $i$-th object can be given by:
$$D_i = \{x_{i1}, x_{i2}, …, x_{iM}\}$$
Again let us assume that the relation $x_i = x_k$ does not mean that $x_i$ and $x_k$ are the same objects in the real world database. It means that the two objects has equal values for the attribute set $A=\{a_1, a_2, a_3, …, a_m\}$. The main objective of the algorithm is to find out the value $k$ automatically in prior to partition the dataset into k disjoint subsets. For distance calculation the distance measure sum of square Euclidian distance is used in this algorithm. It aims at minimizing the average square error criterion which is a good measure of the within cluster variation across all the partitions. Thus the average square error criterion tries to make the $k$-clusters as compact and separated as possible.

Let us assume a set of means $M=\{m_j, j=1, 2, …, K\}$ which consists of initial set of means that has been generated by the algorithm based on the dataset. Based on these initial means the dataset will be grouped into $K$ clusters. Let us assume the set of clusters as $C=(c_j, j=1,2,…,M)$. In the next phase the means has to be updated.

In the algorithm the distance threshold has been taken as:

$$dx = \mu \pm 1\sigma \quad (2)$$

$$\text{where} \quad \mu = \frac{\sum_{i=1}^{n} x_i}{n}$$

$$\text{and} \quad \sigma = \sqrt{\frac{\sum_{i=1}^{n}(x_i - m_0)^2}{n}}$$

Before the searching of initial means the original dataset $D$ will be copied to a temporary dataset $T$. This dataset will be used only in initial set of means generation process. The algorithm will be repeated for n times (where n is the number of objects in the dataset). The algorithm will select the first mean of the initial mean set randomly from the dataset. Then the object selected as mean will be removed from the temporary dataset. The procedure Distance_Threshold will compute the distance threshold as given in eq. 2.

Whenever a new object is considered as the candidate for a cluster mean, its average distance with existing means will be calculated as given in the equation below.

$$\text{Average\_Distance} = \frac{1}{m}\left(\sum_{i=1}^{m} d(M_{am}, x_c)\right) \quad (3)$$

where $M$ is the set of initial means, $i=1, 2, \ldots, m$ and $m \leq n$, $m_c$ is the candidate for new cluster mean.

If it satisfies the distance threshold then it will considered as new mean and will be removed from the temporary dataset. The algorithm is as follows:

Input:
    D = {x$_1$, x$_2$, …, x$_n$}    //set of objects
Output:
    K    // Total number of clusters to be generated
    M = {m$_1$, m$_2$, …, m$_k$}  // The set of initial means
Algorithm:
    Copy D to a temporary dataset T
    Calculate Distance_Threshold on T
    Arbitrarily select x$_i$ as m$_1$
    Insert m$_1$ to M
    Remove x$_i$ from T
    For i= 1 to n-k do    // Check for the next mean
    Begin:
        Arbitrarily select x$_i$ as m$_c$
        Set L=0
        For j= 1 to k do //Calculate the avg. dist
        Begin:
            L= L+ Distance(m$_c$,M[j])
        End
        Average_Distance=L/k
        If Average_Distance ≥
            Distance_Threshold then:
            Remove x$_i$ from T
            Insert m$_c$ to M
            k=k+1
    End

## 3 PERFORMANCE ANALYSIS

The AIM is just the extension of K-means to provide the number of clusters to be generated by the K-means algorithm. It also provides the initial set of means to K-means. Therefore it has been decided to make a comperative analysis of the clustering quality of AIM-K-means with convensional K-means. The main difference between the two algorithms is that in case of AIM-K-means it is not necessary to provide the number of clusters to be generated in prior and for K-means, users have to provide the number of clusters to be generated.

In this evaluation process three datasets have been used. They have been fed to the algorithms according to the increasing order of their size. The programs were developed in C. To test the algorithms thoroughly, separate programs were developed for AIM, AIM-K-means and conventional K-means. In the first phase the datasets have been fed to the K-means with the user fed k-value. Then the AIM-K-means was applied to the same data sets where the value of k means is provided internally by AIM. Lastly the K-means algorithm is again applied to the same datasets with the value of k as given by the AIM-K-means method. The results reavels that:

1. There is a difference in performance for K-means and AIM-K-means algorithm.
2. But the difference reduces when we use K-means algorithm with the value of k as given by the AIM-K-means.

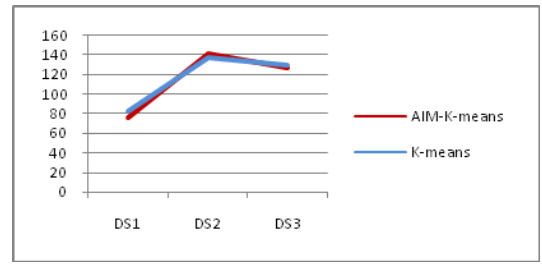

Figure 1: Comparision Based on Average SSE

The above comparison was made on the basis of average sum of square error. From the study it has been found that AIM-K-means is showing improvements in average sum of square. This is basically because of the initial set of cluster means provided to the algorithm. In case of K-means the value of k has been provided based on the output provided by AIM. But it is not possible to provide initial set of clusters in K-means.

## 4 CONCLUSION

The most attractive property of the *K-means* algorithm in data mining is its efficiency in clustering large data sets. But the main disadvantage it is facing is the number of clusters that is to be provided from the user. The algorithm AIM, which is an extension of K-means, can be used to enhance the efficiency automating the selection of the initial means. From the experiments it has been found that it can improve the cluster generation process of the K-means algorithm, without diminishing the clustering quality in most of the cases. The basic idea of AIM is to keep the simplicity and scalability of K-means, while achieving automaticity.

## ACKNOWLEDGMENT

This work has been carried out as part of Research Promotion Scheme (RPS) Project funded by All India Council for Technical Education, Government of India; vide sanction order 8023/BOR/RID/RPS-217/2007-08.

**Samarjeet Borah** has obtained his M. Tech. degree in Information Technology from Tezpur University, India in the year 2006. His major field of study is data mining. He is a faculty member in the Department of Computer Science & Engineering in Sikkim Manipal Institute of Technology, Sikkim, India. He is the principal investigator in a research project sponsored by Government of India. Till date he has published ten papers in various conferences and journals. Borah is a member of the Computer Society of India, International Associations of Engineers, Hong Kong and International Association of Computer Science and Information Technology, Singapore. He has received an award on excellency in research initiatives from Sikkim Manipal University of Health Medical & Technological Sciences.

**Dr. Mrinal Kanti Ghose** has obtained his Ph.D. from Dibrugarh University, Assam, India in 1981. He is currently working as the Professor and Head of the Department of Computer Science & Engineering at Sikkim Manipal Institute of Technology, Mazitar, Sikkim, India. Prior to this, Dr. Ghose worked in the internationally reputed R & D organisation ISRO – during 1981 to 1994 at Vikram Sarabhai Space Centre, ISRO, Trivandrum in the areas of Mission simulation and Quality & Reliability Analysis of ISRO Launch vehicles and Satellite systems and during 1995 to 2006 at Regional Remote Sensing Service Centre, ISRO, IIT Campus, Kharagpur(WB), India in the areas of RS & GIS techniques for the natural resources management. Dr. Ghose has conducted quite a number of Seminars, Workshop and Training programmes in the above areas and published around 35 technical papers in various national and international journals in addition to presentation/ publication of 125 research papers in international/ national conferences. He has guided many M. Tech and Ph.D projects and extended consultancy services to many reputed institutes of the country. Dr. Ghose is the Life Member of Indian Association for Productivity, Quality & Reliability, Kolkata, National Institute of Quality & Reliability, Trivandrum, Society for R & D Managers of India, Trivandrum and Indian Remote Sensing Society, IIRS, Dehradun.